\documentclass[letterpaper,10pt,conference]{ieeeconf}
\IEEEoverridecommandlockouts       
\usepackage{amssymb}
\usepackage{amsmath}
\usepackage[english]{babel}
\usepackage{float}
\usepackage{graphicx}
\usepackage{hyperref}
\usepackage{mathtools}

\usepackage{filecontents}
\usepackage[noadjust]{cite}
\usepackage[font=footnotesize,labelfont=footnotesize]{subcaption}
\usepackage[font=footnotesize,labelfont=footnotesize]{caption}
\usepackage{comment}
\usepackage{authblk}
\usepackage{multirow}
\usepackage{xcolor}
\usepackage{algorithm}
\usepackage{algorithmic}
\usepackage{xspace}
\usepackage{graphicx}
\usepackage{cuted}
\usepackage{capt-of}
\usepackage{siunitx}
\usepackage[normalem]{ulem}
\usepackage{contour}
\usepackage[T1]{fontenc} 

\definecolor{darkblue}{rgb}{0.15,0.15,0.55}
\definecolor{lightgrey}{rgb}{0.75,0.75,0.75}

\contourlength{0.8pt} 

\usepackage{calc}
\newcommand{\ul}[1]{\uline{#1}}

\newcommand{\smallf}[1]{{\small\textsf{#1}}}
\newcommand{\footnotef}[1]{{\footnotesize\textsf{#1}}}

\begin{document}
\title{\LARGE \bf DEX-Mouse: A Low-cost Portable and Universal Interface with Force Feedback for Data Collection of Dexterous Robotic Hands}


\author{Joonho Koh$^{1,\dagger}$, Haechan Jung$^{2,\dagger}$, Nayoung Kim$^2$, Wook Ko$^3$, and Changjoo Nam$^{2,4,*}$\thanks{This work was supported by the National Research Foundation of Korea (NRF) grants funded by the Korea government (MSIT) (No. RS-2022-NR071936 and RS-2024-00461583). $^\dagger$These authors contributed equally to this work. $^1$Dept. of Artificial Intelligence, Sogang University, Korea, $^2$Dept. of Electronic Engineering, Sogang University, Korea, $^2$Dept. of Mechanical Engineering, Sogang University, Korea, $^4$Vertical Labs, Co., Ltd., Korea. $^*$Corresponding author: {\tt\small cjnam@sogang.ac.kr}}}

\maketitle

\begin{strip}
    \vspace{-70pt}
    \centering
    \includegraphics[width=1\textwidth]{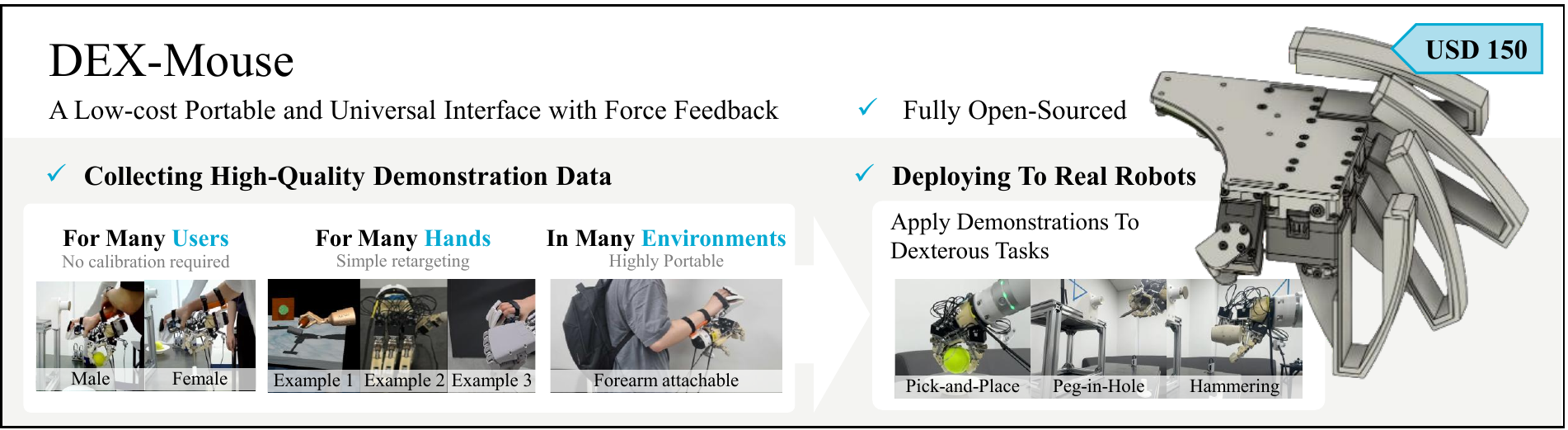}
    \captionof{figure}{An overview of DEX-Mouse. The proposed USD~150, fully open-sourced teleoperation interface enables scalable collection of high-quality demonstration data. Its calibration-free and highly portable design seamlessly accommodates diverse operators, various target robot hands, and diverse real-world environments. Furthermore, collected demonstrations can be directly deployed to diverse dexterous manipulation tasks.}
    \label{fig:teaser}
\end{strip}

\begin{abstract}
Data-driven dexterous hand manipulation requires large-scale, physically consistent demonstration data. Simulation and video-based methods suffer from sim-to-real gaps and retargeting problems, while MoCap glove-based teleoperation systems require per-operator calibration and lack portability, as the robot hand is typically fixed to a stationary arm. Portable alternatives improve mobility but lack cross-platform and cross-operator compatibility. 
We present DEX-Mouse, a portable, calibration-free hand-held teleoperation interface with integrated kinesthetic force feedback, built from commercial off-the-shelf components under USD~150. The operator-agnostic design requires no calibration or structural modification, enabling immediate deployment across diverse environments and platforms. The interface supports a configuration in which the target robot hand is mounted directly on the forearm of an operator, producing robot-aligned data. 
In a comparative user study across various dexterous manipulation tasks, operators using the proposed system achieved an 86.67\% task completion rate under the attached configuration. Also, we found that the attached configuration reduced the perceived workload of the operators compared to spatially separated teleoperation setups across all compared interfaces. The complete hardware and software stack, including bill of materials, CAD models, and firmware, is open-sourced at \url{https://dex-mouse.github.io/} to facilitate replication and adoption.
\end{abstract}

\section{Introduction}
Data-driven approaches such as Imitation Learning (IL) and Vision-Language-Action (VLA) models have shown promise for dexterous hand manipulation. However, collecting high-quality demonstration data at scale remains a key challenge, and the difficulty is compounded by three practical requirements. First, the teleoperation interface must be easy enough to use that human operators can perform demonstrations repeatedly without undue burden~\cite{chi2024universal}. Second, demonstrations must be physically valid with how the target robot moves and interacts, which is difficult to capture from video or simulation alone~\cite{zhao2023learning}. Third, the dataset must have sufficient distributional diversity to support generalization across novel objects and environments~\cite{pmlr-v229-walke23a}.

Simulation-based approaches (e.g., \cite{maddukuri2025sim}) suffer from sim-to-real gaps caused by inaccurate modeling of contact dynamics, friction, and material properties. Video-based approaches (e.g., \cite{hu2024video, ye2025video2policy}) face retargeting problems stemming from morphological differences between the human hand and the target robot. Teleoperation via MoCap gloves (e.g., \cite{zhang2025doglove, shaw2025bimanual}) collects demonstrations on physical robots, ensuring the data captures real contact dynamics and interactions. However, these systems require per-operator calibration. More broadly, conventional teleoperation setups are difficult to relocate. The robot hand, arm, and supporting structures form a bulky integrated system, which strictly limits data collection to fixed environments. To mitigate these mobility and scalability issues, portable hand-held motion-capture interfaces have emerged as a recent alternative (\cite{xu2025dexumi, fang2025dexop}). While these interfaces improve portability, they require per-operator hardware customization due to varying hand sizes and lack compatibility across different robot platforms.

As illustrated in Fig.~\ref{fig:teaser}, we propose DEX-Mouse, a low-cost, calibration-free, and portable hand-held teleoperation interface with integrated force feedback. To improve physical alignment between the operator and the robot hand, our interface supports a configuration in which the target robot hand is mounted directly on the forearm of the operator, while a spatially separated teleoperation setup is also available. Also, its current-based force feedback renders physical resistance for contact-rich manipulation, resulting in improved kinesthetic awareness and more precise force regulation by the operator. This configuration ensures that collected demonstrations directly capture the actual kinematics and dynamics of the target embodiment. Thus, DEX-Mouse produces robot-aligned data through simple proportional retargeting, thereby circumventing complex morphological retargeting. Furthermore, the operator-agnostic design of the interface enables immediate deployment across diverse operators. 

DEX-Mouse offers three key advantages over existing interfaces:
\begin{itemize}
\item \textbf{Universal data-collection system}: An operator- and robot-agnostic teleoperation system accommodates diverse human anatomies and dexterous hands without per-operator calibration or structural modifications, enabling immediate deployment across various platforms.
\item \textbf{Scalable real-robot data collection}: A self-contained, forearm-mounted design enables deployment in arbitrary environments without requiring a stationary robot arm. This portability reduces laborious work to relocate the system, accelerating the acquisition of large-scale demonstrations.
\item \textbf{Low-cost open-sourced device}: The complete hardware and software stack, featuring a USD~150 bill of materials, is open-sourced. This cost-effectiveness lowers the barrier to entry for large-scale dexterous manipulation research, democratizing efficient high-quality data collection.
\end{itemize}

\section{Related Work}

Simulation environments enable large-scale data collection but introduce sim-to-real gaps due to inaccurate modeling of contact dynamics~\cite{maddukuri2025sim}. Video-based approaches extract demonstrations directly from visual observations~\cite{hu2024video, ye2025video2policy, nair2022r3m, ma2022vip, radosavovic2023real}, leveraging real-world data. However, contact-rich manipulation frequently causes occlusion, and translating human motion to robotic systems introduces retargeting errors due to morphological differences.

Teleoperation enables direct transfer of human demonstrations to robotic systems. Existing interfaces include vision-based systems~\cite{Handa2020dexpilot, qin2023anyteleop, yang2024ace}, VR/AR controllers~\cite{cheng2024open, ding2025bunny, iyer2024open}, MoCap gloves~\cite{shaw2025bimanual, wang2024dexcap}, and exoskeletons~\cite{zhang2025doglove, sung2024snu}, which typically retarget human hand motion to the robot via joint-space or vector-based mapping. Vision and VR/AR systems suffer from occlusion during contact-rich tasks, while MoCap gloves and exoskeletons require per-operator calibration and customized manufacturing. More broadly, these hardware-intensive systems involve heavy form factors and restricted mobility, limiting data collection to fixed environments.

Portable hand-held interfaces address the mobility constraints of traditional teleoperation. Operating the end-effector directly eliminates control latency and captures natural physical interactions~\cite{chi2024universal}. Subsequent work extended this paradigm to dexterous manipulation~\cite{xu2025dexumi, fang2025dexop}. However, these interfaces require hardware customization for specific operators and robots, and rely on complex post-hoc visual processing. DEX-Mouse addresses these limitations with an operator-agnostic, cross-embodiment compatible design that operates in real time without per-operator calibration or post-hoc processing.

Across these approaches, a consistent gap remains: no existing system simultaneously achieves portability, operator-agnosticism, cross-embodiment compatibility, and physical consistency of collected data without per-operator calibration or post-hoc processing. DEX-Mouse is designed to address this gap by combining a forearm-mounted, hand-held interface with kinesthetic force feedback in a low-cost, open-sourced package accessible to a broad range of operators and robot platforms.
 
\section{Design Philosophies} 
To overcome the practical bottlenecks characteristic of existing teleoperation systems—namely, limited compatibility across diverse operators and robot platforms, mechanical complexity with high maintenance overhead, and prohibitive equipment costs—we establish three core design principles for DEX-Mouse: universal applicability, mechanical simplicity, and cost-effectiveness.

\subsection{Universal Applicability}
Inspired by the standard computer mouse, which functions identically across different computer architectures without hardware modification, we aim to design a device that is agnostic to both operator anatomy and target robot morphology. The goal is to eliminate dependency on anthropometric calibration, enabling true plug-and-play deployability across diverse operators and robot platforms without structural modifications. (This principle is reflected in the design described in Sec.~\ref{sec:finger_module}, \ref{sec:thumb_module}, \ref{sec:direct_kinematic_mapping} and \ref{sec:portable_sys_integration}.)

\subsection{Mechanical Simplicity}
Highly articulated haptic mechanisms tend to introduce mechanical fragility and maintenance overhead. We adopt a minimalist design philosophy that reduces potential failure points, ensuring the durability required for long-term data collection. Minimal assembly complexity also allows the broader research community to reproduce, modify, and maintain the device without specialized manufacturing equipment. (This principle is reflected in the design described in Sec.~\ref{sec:finger_module}, \ref{sec:thumb_module}, and \ref{sec:mcu_arch}.)

\subsection{Cost-Effectiveness}
High-fidelity motion capture gloves and VR controllers often cost thousands of dollars, limiting the number of devices a research laboratory can deploy and restricting the scale and diversity of data collection efforts. We target a design built entirely from off-the-shelf components with a total bill of materials under USD~150, lowering the financial and technical barriers to entry for the broader research community. (This principle is reflected in the design described in Sec.~\ref{sec:finger_module}, \ref{sec:thumb_module}, and \ref{sec:mcu_arch}.)

\section{Proposed System: DEX-Mouse}

The physical dimensions of the system are derived from the ANSUR II anthropometric dataset~\cite{ansur2} to ensure compatibility across diverse operators. The system employs a modular architecture using off-the-shelf components.

\subsection{Tendon-Driven Finger Module} \label{sec:finger_module}
To teleoperate the fingers (excluding the thumb), we employ the Dynamixel XL330-M077-T smart actuator owing to its minimal footprint ($20 \times 34 \times 26$ mm) and high torque transparency. 

\begin{figure}[t]
\centering
\includegraphics[width=0.4\textwidth]{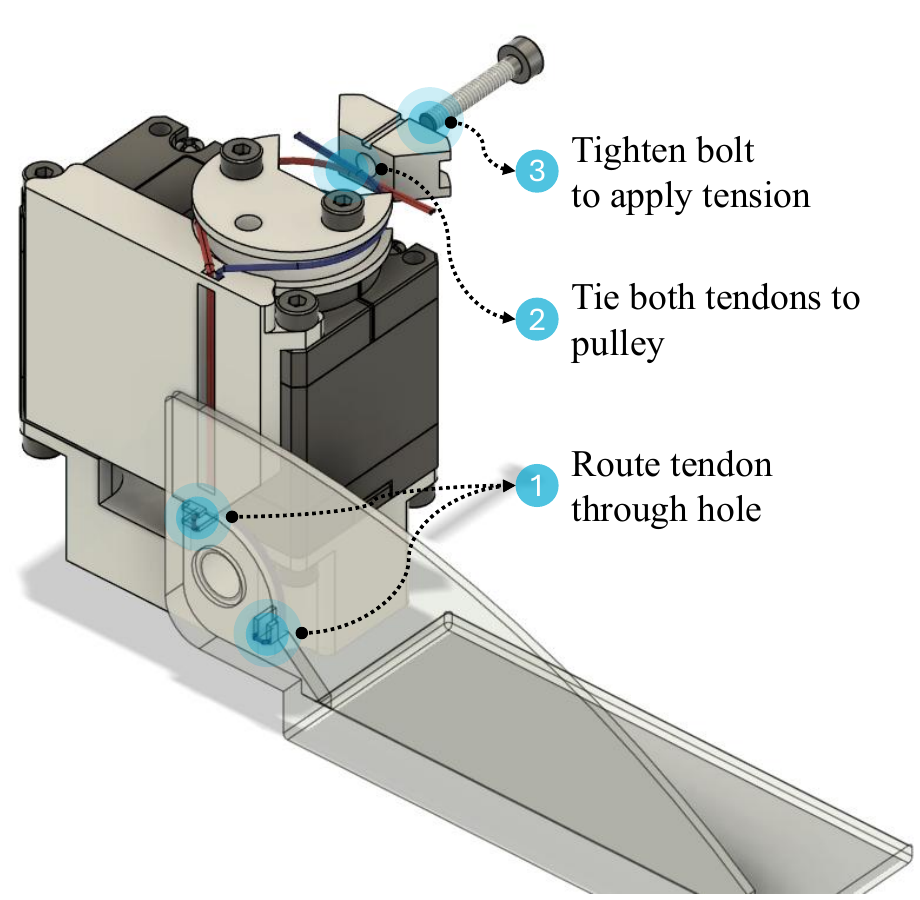}
\caption{Dual tendons facilitate bidirectional finger movement. A single thread is folded and assembled in three steps: (1) routing through guide holes, (2) tying to the pulley, (3) tightening via the tensioning bolt.}
\label{fig:module_design}
\end{figure}

\subsubsection{Anthropometric Compatibility} \label{sec:anthropometric_optimization}
The width of the finger module determines the physical compatibility across operators. According to the ANSUR II database, the 50th percentile hand breadth is approximately 90~mm for males and 78~mm for females. The inclusion of structural housing and manufacturing tolerances results in a module width of 23.5~mm, yielding a total system width of 92.8~mm. This total width exceeds the 50th percentile female hand breadth. We handle this geometric discrepancy by omitting rigid Abduction/Adduction (AA) joints. The absence of lateral kinematic constraints allows operators to naturally splay the fingers, mechanically compensating for the width difference and ensuring usability across a wide range of adult hand sizes.

\subsubsection{Actuation and Transmission} \label{sec:actuation_transmission}
To implement force feedback without expensive torque sensors, maximizing the back-drivability of the system is essential. The M077 variant is selected for its low gear ratio (77.5:1), which provides higher force transparency compared to higher-reduction models. The transmission employs an internal routing architecture that translates the rotational motion of the actuator into the linear displacement of the finger links. A spool is directly coupled to the output shaft of the actuator to drive the tendon through a 1:1 mapping. 

Within the module housing, guide channels define the routing path to prevent derailment. These channels ensure that the high torque transparency of the actuator is maintained throughout the transmission path while keeping the overall footprint compact. For the tendon material, Polyethylene (PE) braided lines are used to minimize elongation under tension and prevent hysteresis, ensuring accurate transmission of the position of the finger to the robot even after prolonged operation. The complex tendon routing mechanism of the module is illustrated in Fig.~\ref{fig:module_design}.

\subsection{Direct-Driven Thumb Module} \label{sec:thumb_module}
Unlike the linear arrangement of the finger modules, the anatomical position of the thumb provides sufficient spatial clearance for a direct actuation mechanism. The Flexion/Extension (FE) joint of the thumb module is therefore directly driven by the actuator, ensuring rigid and immediate torque transmission and improving control bandwidth compared to the tendon-driven counterparts.

To capture the Abduction/Adduction (AA) joint motion---which is critical for dexterous manipulation primitives such as opposition and precision grasping~\cite{cutkosky1989grasp, feix2015grasp}---an AS5600 non-contact magnetic encoder is integrated at the base of the AA joint. This provides high-resolution measurement of the orientation of the thumb for mapping the opposition workspace of the human hand to the robot. Owing to spatial constraints within the palm, haptic feedback is provided for the FE joint only, and the AA joint functions as a passive sensing degree of freedom.

\subsection{Embedded Control Architecture} \label{sec:mcu_arch}
The firmware runs on an STM32F410C8 at a control loop frequency of 100 Hz. Direct Memory Access is used for RS-485 communication with the host PC and UART communication with the actuators to minimize latency. Raw data from the AS5600 magnetic encoder at the AA joint of the thumb is processed via an Exponential Moving Average filter ($\alpha=0.1$) to suppress sensor noise while maintaining responsiveness.

\subsection{Force Feedback Implementation}
The force feedback logic balances transparency during free motion and stiffness during contact using a current-based position control mode, where the actuator acts as a virtual spring.

\subsubsection{Dynamic Gain Scheduling}
The stiffness gain is modulated based on the velocity of the finger of the operator to minimize the feeling of mechanical resistance during free motion.
\begin{equation}
    K_p(v) =
    \begin{cases}
    K_\text{nominal} & \text{if } |v| \le v_\text{th} \text{ (Contact/Static)} \\
    \gamma K_\text{nominal} & \text{if } |v| > v_\text{th} \text{ (Free Motion)}
    \end{cases}
\end{equation}
where $v$ is the finger velocity, $v_\text{th}$ is the speed threshold (set to 20 tick / 10 ms, $\approx$175$^\circ$/s), $K_\text{nominal}=5.0$ is the nominal stiffness gain, and $\gamma \in [0,1]$ is a reduction ratio (set to 0.1 in our firmware). The operator experiences low resistance during rapid movements but feels increased stiffness when grasping an object.


\subsubsection{Unidirectional Force Rendering}
In bilateral teleoperation, the robot can impose reverse forces on the hand of the operator when it retracts or opens, disrupting natural force perception. To prevent this, a unidirectional constraint is implemented where feedback force is generated only when the physical finger of the operator penetrates the virtual finger position of the robot (i.e., when the robot is blocked by an object).
\begin{equation}
    \tau_\text{cmd} =
    \begin{cases}
    K_p(v) \cdot (q_\text{robot} - q_\text{operator}) & \text{if } (q_\text{robot} - q_\text{operator}) > \epsilon \\
    0 & \text{otherwise}
    \end{cases}
\end{equation}
where $\epsilon=100$~tick ($\approx$8.8$^\circ$) is a dead-zone threshold to prevent oscillation. The goal position of the actuator is dynamically updated relative to the current position of the operator, effectively creating a virtual wall that resists further penetration but allows the finger of the operator to withdraw freely.

\subsection{Simple Proportional Retargeting} \label{sec:direct_kinematic_mapping}
DEX-Mouse employs simple proportional retargeting. The structural design of the interface naturally aligns the flexion and extension of the finger of the operator with the corresponding actuation modules. Consequently, the measured actuator positions are directly scaled to the joint limits of the target robot hand using simple linear interpolation. This mapping strategy eliminates the need for per-operator anthropometric calibration, reduces computational latency, and ensures real-time responsiveness during highly dynamic manipulation tasks. The system requires no per-operator calibration. For target hands whose finger FE is driven by a single active DoF, the mapping reduces to direct 1:1 scaling. For hands with multiple active FE DoF per finger, a proportional profile is specified once per target hand and reused across operators.

\subsection{Portable System Integration} \label{sec:portable_sys_integration}
The portable system integration encompasses all components beyond the hand interface: the ergonomic mounting, pose tracker, visual perception unit, and onboard power supply, collectively enabling untethered data collection across diverse environments.

\subsubsection{Ergonomic Mounting Interface} 
The data collection system integrates the DEX-Mouse interface and a target robot hand by an operator holding DEX-Mouse while attaching the hand to his/her forearm. The DEX-Mouse is secured to the palm of the operator via adjustable velcro straps. Concurrently, the target robot hand is mounted directly on the forearm of the operator. Placing the heavy robot hand proximally on the forearm, rather than distally near the wrist, reduces the moment of inertia and gravitational torque acting on the shoulder and elbow joints.

\subsubsection{Pose Tracking} 
Unlike robot arms where end-effector pose can be computed through forward kinematics from joint encoder readings, the forearm-attached nature of DEX-Mouse precludes the use of a known kinematic chain. To obtain the pose of the hand, VIVE Ultimate Tracker~\cite{vive_tracker} is integrated directly onto the device. The tracker employs inside-out tracking technology to estimate global pose independently, allowing the operator to move freely without any external infrastructure.

\subsubsection{Visual Information} 
A Logitech HD Webcam C270 camera is mounted under the target robot hand. This camera captures aligned RGB streams at 30 Hz, providing visual context for training manipulation policies.

\subsubsection{Self-contained Design}
The entire system is untethered, powered by a portable USB-PD battery, with logging handled on a standard laptop; the full kit fits in a single backpack.

\begin{figure}
    \centering
    \includegraphics[width=0.8\linewidth]{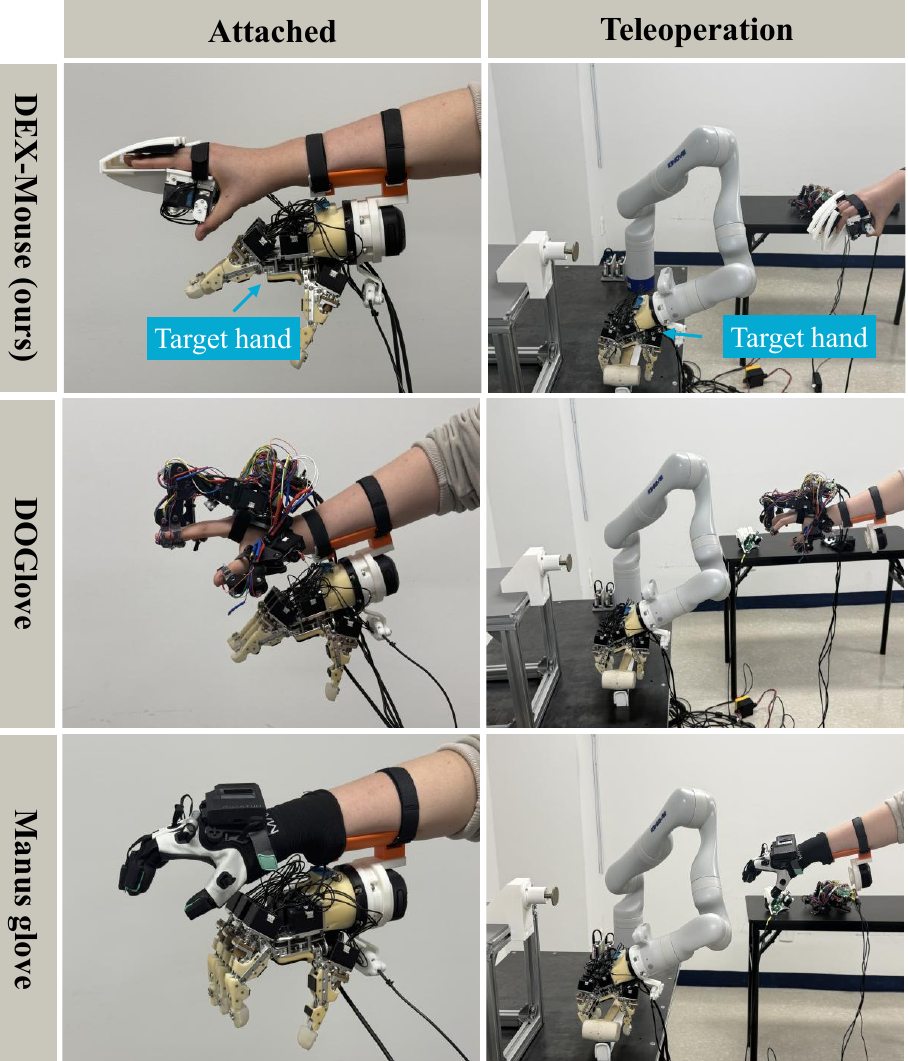}
    \caption{\footnotef{Attached} (left column): The teleoperation interface and the target robot hand are co-located on the arm of the operator. \footnotef{Teleoperation} (right column): The operator wears the interface to control a target robot located in a spatially separated, fixed workspace. The rows correspond to the three different interface devices evaluated.}
    \label{fig:setup_configs}
\end{figure}

\begin{figure*}
    \centering
    \includegraphics[width=1\linewidth]{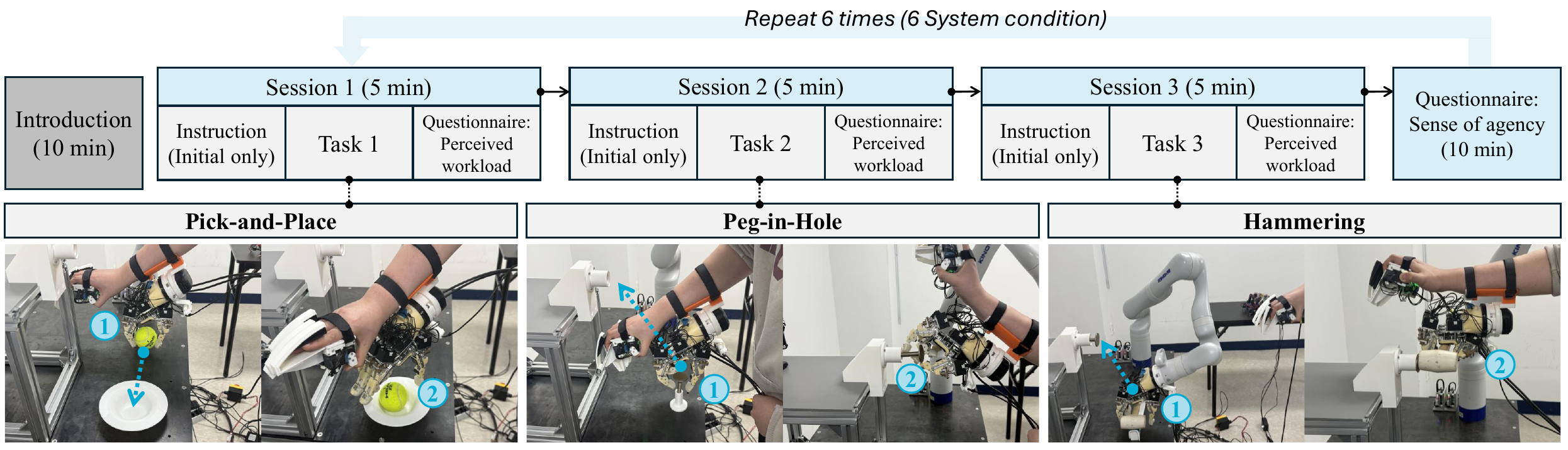}
    \caption{An overview of the experimental procedure and evaluated manipulation tasks. The study follows a $3 \times 2$ within-subjects design across six randomized conditions. The procedural workflow (top) illustrates the sequence for each condition: participants complete one practice trial and five main evaluation trials per task, followed immediately by a perceived workload assessment. A SoA questionnaire concludes each condition. The evaluated tasks (bottom) include Pick-and-Place for basic reach-and-grasp performance, Peg-in-Hole for precise visuomotor coordination and fine force control, and Hammering for grasp stability under dynamic impacts.}
    \label{fig:procedure}
\end{figure*}

\section{Experiments}
To quantitatively evaluate the proposed interface beyond qualitative demonstrations, we conduct a comparative user study against state-of-the-art devices. To address practical requirements for scalable data collection, we measure task success rate and completion time to verify the efficiency of data collection, alongside perceived workload to ensure minimal operator fatigue. Finally, we train a policy using the collected demonstrations to validate downstream learning utility. 
The study was approved by the Sogang University Institutional Review Board (IRB: SGUIRB-A-2501-04-2).

\subsection{Experimental Setup}

\subsubsection{Participants}
A total of eight participants over the age of 19 participated in the study, including four males ($\mathit{M} = 27.25$, $\mathit{SD} = 3.96$) and four females ($\mathit{M} = 23.75$, $\mathit{SD} = 1.92$).  All participants received $12$ US dollars as compensation.

\subsubsection{Experimental Design}
Our study employed a $3 \times 2$ within-subjects design to examine the effects of interface type and teleoperation configuration. The first factor was \ul{Interface Type}, which included \smallf{DEX-Mouse} (proposed), \smallf{DOGlove}~\cite{zhang2025doglove}, and the \smallf{Manus Quantum Metagloves} (hereafter, \smallf{Manus glove})~\cite{manus_quantum}. The second factor was \ul{Collection Configuration}, which included \smallf{Attached}, where the interface and the robot-hand tracking frame were physically co-located on the arm of the operator, and \smallf{Teleoperation}, where the interface was worn but the robot operated in a spatially separated fixed workspace (Fig.~\ref{fig:setup_configs}). Each participant therefore experienced six data-collection conditions in total.

Across all experimental conditions, we used a four-fingered Blue Robin dexterous hand~\cite{bluerobin_hand}, where each finger is driven by two active joints. The fingers also employ a spring-loaded multi four-bar linkage mechanism. The design of DEX-Mouse is not specific to this robot hand. The current implementation uses it solely due to hardware availability.
For \smallf{Teleoperation}, a Kinova Gen3 7-degree-of-freedom (DoF) manipulator served as the follower arm where the robotic hand was attached.
End-effector commands from VIVE Ultimate Tracker were resolved into joint targets via Pinocchio inverse kinematics solver~\cite{carpentier2019pinocchio} at 20\,Hz. 
In \smallf{Attached}, the target robot hand was directly mounted on the forearm of the operator.

\subsubsection{Experimental Procedure}
Participants were first briefed about the study and provided informed consent. They then experienced six system conditions in a randomized order, following the experimental procedure detailed in Fig.~\ref{fig:procedure}. For conditions utilizing the DOGlove and Manus glove, a per-operator calibration was performed prior to the trials. In the first system condition, participants watched an instructional video that explained how to perform the tasks. They then completed one practice trial and five main trials for each task. In the remaining system conditions, participants completed the same sequence of tasks with one practice trial followed by five main trials for each task, without additional instructions. We instructed all participants to complete each task as quickly and accurately as possible.

Immediately after completing the five main trials for a given task, participants assessed their perceived workload using NASA Raw Task Load Index (NASA-RTLX)~\cite{hart2006nasa}. Once all 15 main trials for a system condition were finished, participants completed a Sense of Agency (SoA) questionnaire~\cite{tapal2017sense} during a rest period before moving to the next condition. This questionnaire evaluates the subjective feeling of controlling the actions of the robot.

Participants completed three tasks in a fixed order across all system conditions: pick-and-place, peg-in-hole, and hammering (Fig.~\ref{fig:procedure}). These tasks were strategically selected to evaluate a comprehensive range of dexterous manipulation capabilities, encompassing distinct grasp taxonomies and dynamic force requirements.
    
\begin{itemize}
    \item \textbf{Pick-and-Place}: Grasp a spherical ball and transfer it to a designated receptacle. This task evaluated basic reach-and-grasp performance and spatial positioning. A trial was marked successful if the ball was stably deposited inside the plate.
    \item \textbf{Peg-in-Hole}: Grasp a nail using a precision pinch grip and insert it into a fitted hole. This task assessed fine visuomotor coordination and multi-finger force regulation. A trial was considered successful when the nail was inserted into the hole.
    \item \textbf{Hammering}: Grasp a hammer using a stable power grasp and strike a pre-positioned nail. This task evaluated grasp stability and interface transparency under dynamic impacts. Success was defined as driving the nail flush with the surface while maintaining a stable, slip-free grasp on the hammer.
\end{itemize}

\subsubsection{Measures}
We evaluated system performance using objective and subjective measures. The objective measures were task success rate across the five main trials for each task and task completion time (recorded only for successful trials). The subjective measures were perceived workload, assessed using NASA-RTLX (six items, Cronbach's $\alpha = .83$), and SoA (11 items, Cronbach's $\alpha = .67$). All items were assessed using a 7-point Likert scale (1 = strongly disagree, 7 = strongly agree).

\begin{table*}[t]
\caption{Detailed Task Performance ($N=8$). Success Rate (SR) is measured in \%, and Completion Time (Time) in seconds. All values are presented as Mean (SD). The bottom rows present the overall average of each configuration across all interfaces.}
\label{tab:detailed_performance}
\centering
\resizebox{\textwidth}{!}{%
\begin{tabular}{|l|c||c|c||c|c||c|c||c|c|}
\hline
\multirow{2}{*}{\textbf{Interface Type}} & \textbf{Collection} & \multicolumn{2}{c||}{\textbf{Pick \& Place}} & \multicolumn{2}{c||}{\textbf{Peg-in-Hole}} & \multicolumn{2}{c||}{\textbf{Hammering}} & \multicolumn{2}{c|}{\textbf{Overall Average}} \\
\cline{3-10}
& \textbf{Configuration} & \textbf{SR (SD)} & \textbf{Time (SD)} & \textbf{SR (SD)} & \textbf{Time (SD)} & \textbf{SR (SD)} & \textbf{Time (SD)} & \textbf{SR (SD)} & \textbf{Time (SD)} \\
\hline
\multirow{2}{*}{\textbf{\footnotef{DEX-Mouse} (Ours)}} 
& \footnotef{Attached} & 95.0 (13.23) & \textbf{5.57 (2.46)} & \textbf{72.5 (28.17)} & \textbf{14.29 (6.41)} & \textbf{92.5 (13.92)} & \textbf{10.29 (3.86)} & \textbf{86.67 (22.11)} & \textbf{10.05 (5.78)} \\
& \footnotef{Teleoperation} & 80.0 (20.00) & 17.80 (4.98) & 17.5 (27.27) & 25.21 (3.87) & 60.0 (31.62) & 17.13 (7.37) & 52.5 (37.33) & 18.77 (6.56) \\
\hline
\multirow{2}{*}{\footnotef{DOGlove}}
& \footnotef{Attached} & \textbf{100.0 (0.00)} & 7.53 (2.23) & 60.0 (30.00) & 16.19 (5.79) & 72.5 (33.07) & 11.24 (3.31) & 77.5 (30.72) & 11.67 (5.47) \\
& \footnotef{Teleoperation} & 80.0 (24.49) & 19.79 (6.24) & 12.5 (13.92) & 27.44 (6.97) & 57.5 (32.31) & 22.73 (6.91) & 50.0 (37.42) & 22.48 (7.24) \\
\hline
\multirow{2}{*}{\footnotef{Manus glove}} 
& \footnotef{Attached} & 92.5 (13.92) & 7.97 (3.59) & 32.5 (26.34) & 17.22 (4.84) & 62.5 (29.05) & 11.93 (4.34) & 62.5 (34.31) & 11.93 (5.59) \\
& \footnotef{Teleoperation} & 62.5 (33.82) & 14.59 (3.25) & 2.5 (6.61) & 30.00 (0.00) & 45.0 (23.98) & 17.93 (4.77) & 36.67 (34.96) & 17.01 (5.39) \\
\hline\hline
\multirow{2}{*}{\textbf{Configuration Average}} 
& \textbf{\footnotef{Attached}} & \textbf{95.83 (11.52)} & \textbf{7.02 (3.01)} & \textbf{55.00 (32.79)} & \textbf{15.90 (5.91)} & \textbf{75.83 (29.43)} & \textbf{11.15 (3.94)} & \textbf{75.56 (31.13)} & \textbf{11.22 (5.68)} \\
& \footnotef{Teleoperation} & 74.17 (27.98) & 17.39 (5.42) & 10.83 (19.13) & 27.55 (5.68) & 54.17 (30.27) & 19.26 (6.91) & 46.39 (37.24) & 19.42 (6.89) \\
\hline
\end{tabular}%
}
\end{table*}

\subsection{Results and Findings}
We evaluated six conditions defined by collection configuration and interface type across the three tasks. For each task, we measured task success rate and task completion time. We also assessed perceived workload and SoA.

\subsubsection{Impact of Collection Configuration}
The physical setup consistently affected teleoperation performance. As summarized in the bottom rows of Table~\ref{tab:detailed_performance}, \smallf{Attached} yielded a higher overall success rate and a faster completion time compared to \smallf{Teleoperation}.

This performance gap was observed across all manipulation tasks. In the pick-and-place task, \smallf{Attached} demonstrated higher spatial positioning efficiency. This performance difference became more pronounced in tasks requiring precise physical interactions. In the peg-in-hole task, \smallf{Teleoperation} resulted in lower success rates, likely due to the mental coordinate transformations required between the interface of the operator and the remote robotic workspace. In contrast, \smallf{Attached} mitigated this spatial discrepancy, improving force regulation. Similarly, in the hammering task, \smallf{Attached} outperformed the remote setup. These results indicate that co-locating the robotic workspace with the proprioception of the operator enables reflexive motor responses and reduces the workload of the operator.

\subsubsection{Impact of Interface Type}
\begin{figure}
    \centering
    \includegraphics[width=0.9\linewidth]{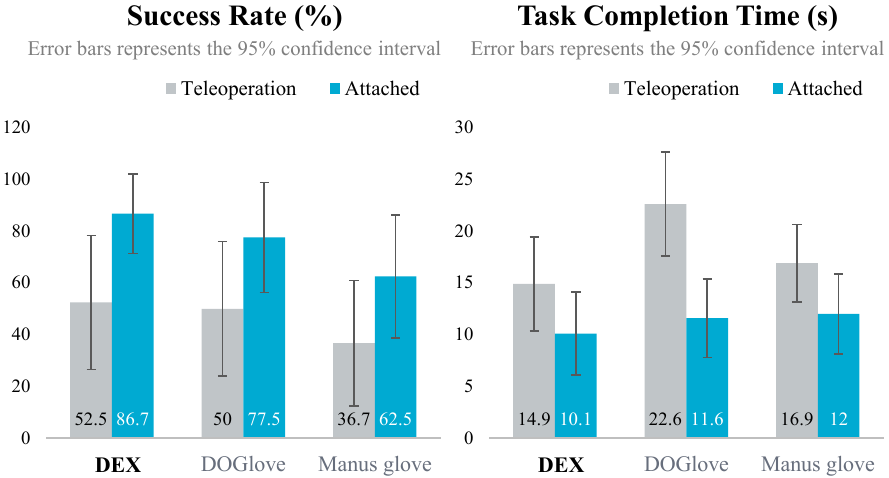}
    \caption{Overall task success rates (\%) and completion times (s) across different interfaces and collection configurations.}
    \label{fig:experiments_result}
\end{figure}

As shown in Table~\ref{tab:detailed_performance} and Fig.~\ref{fig:experiments_result} (focusing on \smallf{Attached}, which outperformed \smallf{Teleoperation}), DEX-Mouse achieved the highest overall success rate and fastest completion time among the three tested interfaces. This performance was achieved with a total component cost of under USD~150, compared to the commercial Manus glove (retail price $\approx$ USD~7,000) and the research-grade DOGlove (manufacture price $\approx$ USD~600).

In the pick-and-place task, all three interfaces achieved high success rates, indicating sufficient visual-motor mapping for basic reaching. However, the performance gap widened during contact-rich tasks. In the peg-in-hole task, the Manus glove resulted in lower success rates, likely due to the absence of kinesthetic feedback. While DOGlove provides haptic feedback, its full high DoF tracking occasionally induced unintended finger micro-motions. In contrast, the structurally constrained 6-DoF design of DEX-Mouse effectively suppressed unnecessary movements, resulting in a higher success rate. Per-participant inspection further revealed that most of the within-cell variance in peg-in-hole performance across all three interfaces originated from two participants who self-reported difficulty with precise thumb-index opposition due to their short thumb length. In the hammering task, the simplified structure of DEX-Mouse enabled the operator to concentrate on thumb engagement for a secure power grasp, while the kinesthetic force feedback provided physical resistance to prevent the tool from slipping.

\begin{figure}
    \centering
    \includegraphics[width=0.7\linewidth]{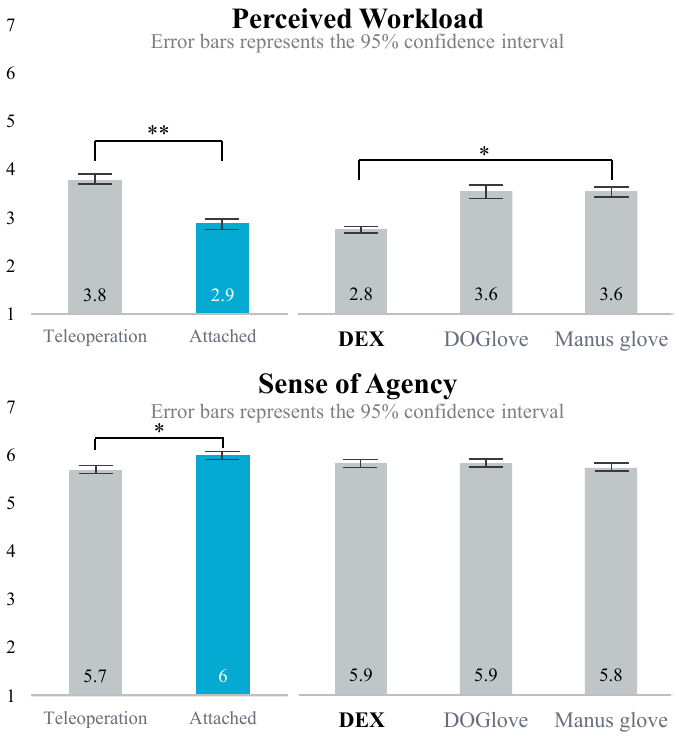}
    \caption{Subjective operator experience results evaluating perceived workload and SoA. * $p < 0.05$, ** $p < 0.01$, *** $p < 0.001$.}
    \label{fig:result_user}
\end{figure}

\subsubsection{Operator Experience}
Fig.~\ref{fig:result_user} illustrates the subjective questionnaire results for perceived workload and SoA. For perceived workload, we found significant main effects of collection configuration [$F(1, 7) = 24.453, p = 0.002, \eta_p^2 = 0.777$] and interface type [$F(2, 14) = 7.720, p = 0.005, \eta_p^2 = 0.524$]. The interaction between collection configuration and interface type was not significant [$F(2, 14) = 1.952, p = 0.218, \eta_p^2 = 0.218$]. Bonferroni post hoc tests showed that perceived workload was significantly lower in the Attached condition than in the Teleoperation condition (95\% CI [0.438, 1.242], $p < 0.01$). In addition, perceived workload was significantly lower with DEX-Mouse than with the Manus glove (95\% CI [0.085, 0.360], $p < 0.05$). No significant differences were found for the other pairwise comparisons.

For SoA, we found a significant main effect of collection configuration [$F(1, 7) = 6.881, p = 0.034, \eta_p^2 = 0.496$]. The main effect of interface type was not significant [$F(2, 14) = 0.588, p = 0.569, \eta_p^2 = 0.077$], and the interaction between collection configuration and interface type was also not significant [$F(2, 14) = 0.145, p = 0.866, \eta_p^2 = 0.020$]. Bonferroni post hoc tests showed that SoA was significantly lower in the Teleoperation condition than in the Attached condition (95\% CI [0.025, 0.490], $p < 0.05$). No other pairwise comparisons showed significant differences.

\subsection{Cross-Embodiment Compatibility}
\begin{figure}
    \centering
    \includegraphics[width=1.0\linewidth]{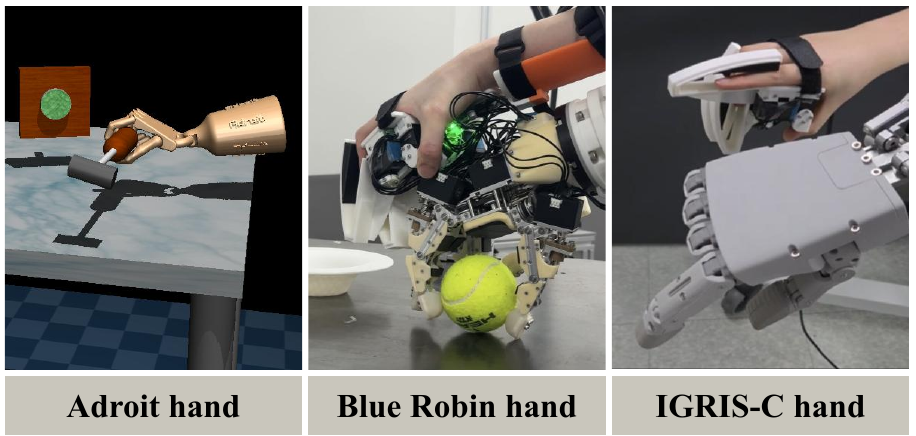}
    \caption{Demonstration of cross-embodiment compatibility. By employing simple proportional retargeting, DEX-Mouse enables calibration-free teleoperation across distinct robot morphologies: (left) a simulated 5-finger active 30-DoF Adroit hand, (center) a 4-finger 8-DoF Blue Robin hand, and (right) a 5-finger 11-DoF (active 6-DoF) humanoid hand (IGRIS-C).}
    \label{fig:cross_embodiment}
\end{figure}

To validate cross-embodiment compatibility, we deployed DEX-Mouse on a simulated 5-finger 30-DoF Adroit hand~\cite{rajeswaran2017learning} and a physical 5-finger 11-DoF humanoid hand~\cite{robros_igrisc}. As illustrated in Fig.~\ref{fig:cross_embodiment}, using simple proportional retargeting without per-operator calibration or structural modifications, an operator successfully performed complex multi-finger manipulations on both platforms. These qualitative results demonstrate the universal applicability of our interface.

\subsection{Downstream Policy Learning}
To verify that demonstrations collected with DEX-Mouse are suitable for policy training, we trained a diffusion policy~\cite{xu2025dexumi} across the three evaluated tasks: pick-and-place, peg-in-hole, and hammering. For each task, we collected 200 demonstrations under \smallf{Attached} configuration. The data collection process required approximately 1--1.5 hours per task, demonstrating that hundreds of demonstrations can be collected efficiently.

The policy architecture follows the DexUMI framework with two modifications. First, while DexUMI relies on relative positional representations (mainly due to the differences between DexUMI and the target hand), our policy uses absolute joint positions for the robot hand. This absolute mapping is feasible because DEX-Mouse captures demonstrations directly on the target embodiment, inherently eliminating the morphological gap. Second, instead of using external tactile sensors, we incorporated the measured motor torque values as proprioceptive inputs to represent contact interactions. 

To evaluate the robustness of the learned policy, we attached the hand to an FR5~\cite{fairino_fr5} cobot to execute tasks. We measured the task success rate over 20 trials for each task. During these trials, the initial poses of the objects were randomized to assess spatial generalization. For the pick-and-place task, the positions of the objects were randomly varied within a $10 \times 10$~cm workspace around their nominal initial states, preserving the relative spatial arrangement between the ball and the receptacle. For the peg-in-hole task, the initial position of the nail was randomized within a $5 \times 5$~cm region, while the hole remained fixed. Similarly, for the hammering task, the target objects remained fixed, whereas the initial position of the robot hand was randomized within a $10 \times 10 \times 10$~cm volume. Under these randomized conditions, the learned policy achieved success rates of 90\% for the pick-and-place task, 50\% for the peg-in-hole task, and 95\% for the hammering task. These results demonstrate that the proposed interface provides robust, robot-aligned data for downstream imitation learning. The 50\% success rate in the peg-in-hole task reflects the compound difficulty of executing a precise pinch grasp followed by a millimeter-tolerance insertion. Under initial spatial randomization, successfully chaining these sequential precision tasks relying solely on a single camera and proprioceptive torque feedback remains inherently challenging.

\section{Limitations and Future Work}

While DEX-Mouse demonstrates high task success rates in dexterous teleoperation, future work will focus on three key areas to further enhance system capabilities and scalability.

First, the current calibration-free design omits rigid AA joints to ensure an operator-agnostic fit. To capture lateral finger motion during tasks requiring substantial finger splay without compromising this universality, we plan to incorporate passive flexure sensors. This addition aims to further improve the kinematic precision of the interface.

Second, while the forearm-mounted configuration mitigates wrist load, long-duration data collection requires further strain reduction. Future iterations will explore lightweight composite materials and active gravity compensation to improve wearability.

Finally, the variance observed in peg-in-hole performance was primarily driven by two participants who self-reported difficulty with precise thumb-index opposition due to short thumb length. Motivated by this observation, future iterations will incorporate an adjustable-length thumb module. The limited sample size (N = 8) also renders formal assumption checks underpowered, so reported effect sizes should be interpreted as upper bounds.

\section{Conclusion}

This paper presented DEX-Mouse, a low-cost, portable, and universal interface designed for dexterous manipulation data collection. By abstracting the kinematics of the human hand into a normalized tendon-driven mechanism, the system achieves a calibration-free architecture that seamlessly accommodates diverse human anatomies and robot morphologies. The integration of sensor-less kinesthetic force feedback and a forearm-mounted configuration significantly improved task success rates and reduced operator workload during contact-rich manipulations. Empirical evaluations showed that DEX-Mouse attains task performance competitive with significantly more expensive commercial MoCap gloves.

Ultimately, DEX-Mouse lowers the technical and financial barriers to entry for dexterous manipulation research. With an open-source design and a total component cost under USD~150, the proposed interface enables the scalable collection of high-fidelity, robot-aligned demonstration data. This accessibility will facilitate the rapid advancement of data-driven robotics across diverse, real-world environments.

\bibliographystyle{IEEEtran}
\bibliography{references}

\end{document}